%% file: main.tex
\documentclass[runningheads]{llncs}
\usepackage{graphicx}
\usepackage{caption}
\usepackage{graphicx}
\usepackage{wrapfig}
\usepackage{xcolor}

\usepackage{textcomp}
\usepackage{graphicx}
\usepackage{amsmath,amssymb}

\usepackage{xspace}

\usepackage{array}
\newcolumntype{R}[1]{>{\raggedleft\let\newline\\\arraybackslash\hspace{0pt}}m{#1}}
\usepackage{arydshln}
\usepackage[utf8]{inputenc}
\usepackage{mathtools}
\makeatletter
\def\blfootnote{\xdef\@thefnmark{}\@footnotetext}
\makeatother

\makeatletter
\def\thfootnote{\xdef\@thefnmark{}\@footnotetext}
\makeatother
\usepackage{booktabs}
\usepackage{multicol}
\usepackage{diagbox}
\usepackage[nolist,nohyperlinks]{acronym}

\usepackage{siunitx}
\usepackage{multirow}

\newcommand{\printfnsymbol}[1]{%
  \textsuperscript{\@*}%
}

\begin{document}
\title{Deep Learning Under the Microscope: Improving the Interpretability of Medical Imaging Neural Networks}
\titlerunning{Deep Learning Under the Microscope}
%


\author{Magdalini Paschali\inst{1}\thanks{The authors contributed equally.} \and Muhammad Ferjad Naeem \inst{1}\printfnsymbol{1} \and Walter Simson\inst{1} \and \\ Katja Steiger\inst{2} \and Martin Mollenhauer\inst{2} \and Nassir Navab\inst{1,3}}
\authorrunning{M. Paschali, M.F. Naeem et al.} 
%
\institute{
Computer Aided Medical Procedures, Technische Universit\"{a}t M\"{u}nchen, Germany
\and 
    Institut f\"{u}r Pathologie,
    Fakult\"{a}t f\"{u}r Medizin,
    Technische Universit\"{a}t M\"{u}nchen, Germany
    \and
Computer Aided Medical Procedures, Johns Hopkins University, USA}

\maketitle              
\begin{abstract}

In this paper, we propose a novel interpretation method tailored to histological Whole Slide Image (WSI) processing. 
A Deep Neural Network (DNN), inspired by Bag-of-Features models is equipped with a Multiple Instance Learning (MIL) branch and trained with weak supervision for WSI classification. 
MIL avoids label ambiguity and enhances our model's expressive power without guiding its attention. We utilize a fine-grained logit heatmap of the models activations to interpret its decision-making process. The proposed method is quantitatively and qualitatively evaluated on two challenging histology datasets, outperforming a variety of baselines. In addition, two expert pathologists were consulted regarding the interpretability provided by our method and acknowledged its potential for integration into several clinical applications.

\end{abstract}
\input{1_introduction}
\input{2_method}
\input{3_experimental_settings}

\input{4_results}
\input{5_conclusion}

%
%
\bibliographystyle{unsrt} 
\bibliography{refs.bib}

\end{document}

%% file: 1_introduction.tex
\section{Introduction}



Deep Neural Networks (DNNs) owe their impressive ability to solve challenging tasks such as classification, segmentation and localization to their complex structure. 
Although inspired by the human brain neurons, DNNs' decision-making process significantly differs from the one of humans. This property can be beneficial in the identification of underlying correlations and features that might be missed by the human eye. However, it could also lead to a decision-making process based on a false premise, which is highly undesirable in critical applications such as medical imaging.
Developing interpretable DNNs that provide comprehensive explanations for their decisions would enable their full integration to Computer Aided Diagnosis (CAD) Systems to assist physicians and alleviate the risk of basing decisions on non-representative image features. 

Recently, a plethora of model interpretation methods have been proposed~\cite{survey_interpretability} with the aim of deciphering DNNs' verdicts by providing human-understandable explanations~\cite{beenkim}.
Several visualization~\cite{saliencymaps_journals/corr/SimonyanVZ13},~\cite{classactivationmaps_zhou2016cvpr},~\cite{guidedattention} and captioning techniques~\cite{captioningsurvey} aim to improve the interpretability of trained models. However, setbacks are preventing those methods from being fully integrated into medical imaging frameworks. Challenging tasks such as histology image analysis require a model to not only provide fine-grained, sub-cellular details in the visualizations, but also to maintain these properties on large Whole Slide Images (WSI). Therefore classical visualization approaches, such as~\cite{classactivationmaps_zhou2016cvpr} are insufficient to interpret such images effectively. 


\begin{figure}[t]
\centering
  \includegraphics[width=\textwidth]{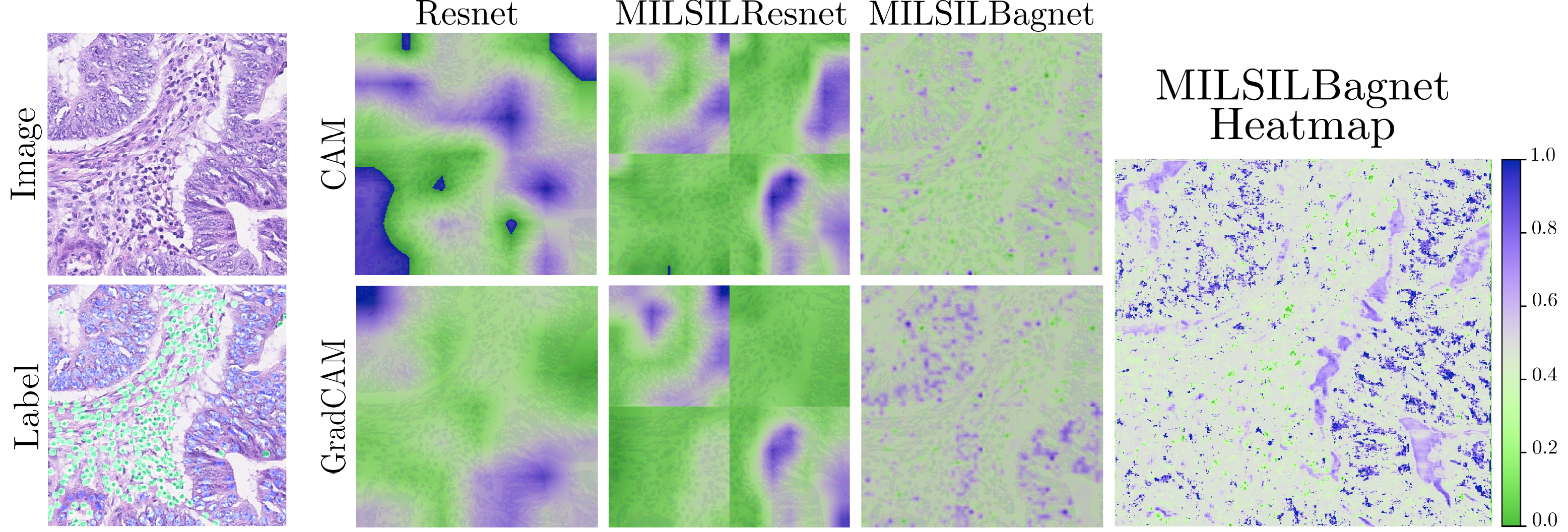}
   \caption{Comparison of the proposed visualization (MILSILBagnet) with traditional techniques~\cite{selvaraju2017grad}. Our model establishes a more direct correspondence between the model's activations and the image allowing for detailed visualization of sub-cellular structures. On the Label, healthy cells are highlighted with green color and malignant with blue.}
   \label{fig:crccompare}
\end{figure}

To this end, a weakly supervised DNN interpretation mechanism is proposed that has been highly specialized to WSI Processing. Our novel approach is tailored towards the interpretation of medical imaging DNNs, combining three specific components that yield visualizations that are fine-grained and scalable: 

1) \textit{Multiple Instance Learning (MIL)}~\cite{milsurvey} is a suitable way to model medical tasks, where only weak supervision is provided as pathologies co-exist within an image and examinations are acquired from multiple views. In a MIL setting samples sharing the same label are grouped into sets referred to as bags. Previous attempts have been made to incorporate MIL in CADs~\cite{milhashing} for medical image retrieval, without predicting the significance of each sample in a bag. Several methods~\cite{icmlmil} aim to learn the importance of each sample within a bag and provide visualizations interpreting the predictions of a model. However, for the use case of WSI, these methods are limited by the need for extensive nuclei-level annotations.

2) \textit{Decisions based majorly on local features within an image}~\cite{bagnet}. In tasks, such as fine-grained classification of WSI,
global awareness of an image can lead to ambiguity as the class label depends mainly on a few nuclei from the entire image.
Our method approximates the classic Bag-of-Features model with a Multiple Instance Learning (MIL) enabled DNN. Combining local features with MIL in a model circumvents the ambiguity induced by the weak annotations.

3) \textit{Fine-grained pixel-level visualization of the attention of DNNs.} The proposed architecture provides highly interpretable heat maps as shown in Fig.~\ref{fig:crccompare}.
These heatmaps can additionally be leveraged by physicians as an initial proposal to increase the annotation speed of WSI.
Speeding up the annotation process would significantly benefit CAD systems, as it would increase the amount of annotated datasets utilized for research. Overall, our contributions can be summarized as follows: 1) We incorporate a MIL branch into a Bag-of-Features inspired model trained with weak supervision.
2) We utilize a highly interpretable logit heatmap as a visualization method for the model's decision process.
3) We perform thorough quantitative and qualitative evaluation of our method on two publicly available datasets. 

For the scope of the paper, the concept of interpretability refers to the correspondence between the model activations and the ground truth.

%% file: 2_method.tex
\begin{figure}[t]
    \centering{   
    \includegraphics[width=\textwidth]{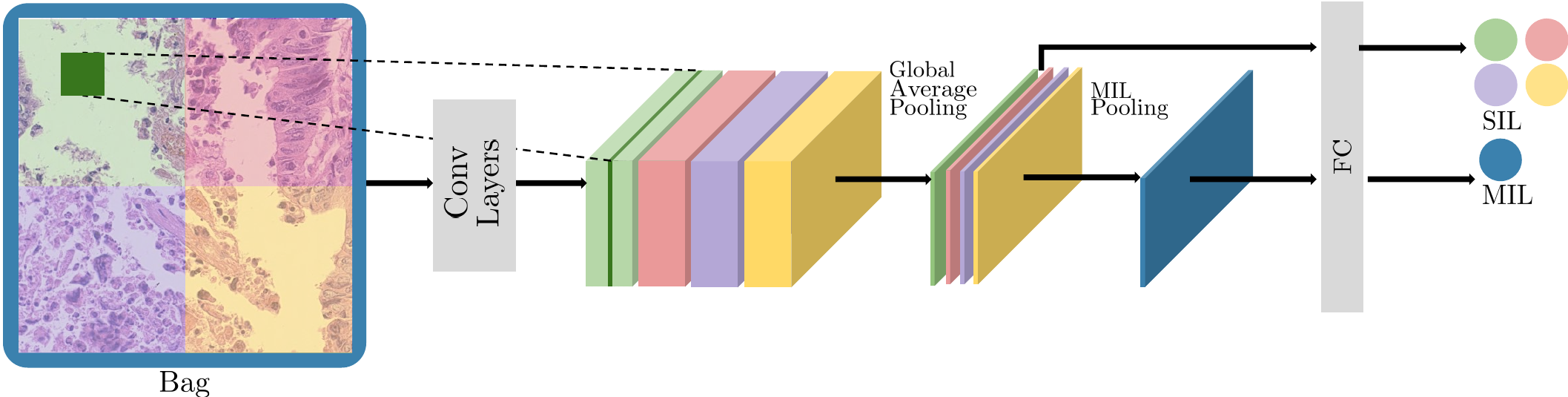}
    \caption{
    Model Overview. 
    The image is processed by the network as a bag of patches. Both the MIL and SIL branches are followed by a linear classifier that outputs patch-level and bag-level labels.
    Every feature vector after the convolutional layers directly corresponds to a $9\times9$ region from the original image.}
    \label{fig:architecture}
    }
\end{figure}

\section{Method}
\subsubsection{Decisions based on local features}
Our main goal, learning to classify a WSI based on local information while leveraging weak labels, can be achieved by expressing the local features within an image as a Bag-of-Features (BoF) ~\cite{nowak2006sampling}. 
The features are processed independently by a linear classifier allowing the decision process to be traced back to individual patches increasing interpretability.

Our method extends the classic BoF approach to DNNs. Built upon BagNet~\cite{bagnet}, the architecture is inspired by ResNet50, where the majority of 3x3 convolutions in the architecture are replaced by 1x1 convolutions, as explained in~\cite{bagnet}. 
This allows every 2048 dimension feature vector after the convolutional blocks to directly correspond to a 9x9 region in an input image.
A spatial global average pooling is performed on these features to compute an image-level feature vector. A linear classifier, in the form of a Fully Connected Layer, performs classification on this vector to infer the logit values, which are afterwards converted to probabilities through a Softmax function.


\noindent\textbf{Multiple Instance Learning BagNet}
Our main contribution regarding the model architecture is the incorporation of a MIL Branch into BagNet. Large histology images are cropped exhaustively into patches to form a bag. Utilizing MIL is suitable for this task, since resizing the original images to fit the input of the model would lead to significant loss of resolution. Furthermore, since we are only leveraging image-level labels for each cropped patch, organizing them into a bag with a single label avoids label ambiguity.

The MIL branch performs an average pooling along the feature dimension of the vectors that belong to the same bag and leads to a vector representing the entire bag. 
Training the model (MILBagnet) exclusively using bag-level labels can lead to sparsity in the gradients~\cite{milhashing}.
Therefore, the final proposed architecture is equipped with both a Single Instance Learning (SIL) and a MIL Branch. After a bag is forwarded into the network, we acquire the SIL feature vectors and the MIL bag-level feature vector. Both these features are processed by the Fully Connected (FC) layer to infer the MIL and SIL labels, as can be seen in Fig.~\ref{fig:architecture}.

Since we jointly train the SIL and MIL branches, the FC layer is optimized for the highest activations in both of them. Hence, the choice of MIL pooling operation should prevent a discrepancy between the activation values. This can be achieved by an average pooling operation, which additionally allows the network to process inputs of different sizes.

\noindent\textbf{Logit Heatmaps}
The proposed visualization method paired with the aforementioned model is class-wise logit heatmaps. The limited receptive field of our DNN, in combination with the global average pooling operations, allows for images of any size multiple of $9\times 9$ to be forwarded into the model. The MIL branch is switched off and a sliding window approach is followed, where a window of size  $9\times9$ is placed around every single pixel of the image. The logit value for the window is calculated and placed at the respective pixel position in the per class heatmap.
Since the logit values are derived from such small regions of the image, the pixels with the highest activations are representative of the model's attention. 
Observing the attention map, we can better understand the model's decision-making process.
Especially in the context of MIL, it allows us to interpret the importance of each section of the image to a global decision.

%% file: 3_experimental_settings.tex
\section{Experimental Settings}

\noindent\textbf{Datasets}
\label{sec:dataset}
The proposed method was evaluated on two publicly available histology datasets: HistoPhenotypes (CRC)~\cite{crchistophenotypes} and CAMELYON16 (CAM16)~\cite{camelyon16}. CRC consists of one hundred patches extracted from 10 H\&E (Haemotoxylin and Eosin) stained WSI from 9 patients with colorectal adenocarcinomas. The nuclei were annotated extensively (but not completely) into epithelial, inflammatory, fibroblast, and miscellaneous. For our experiments, we classified the patches as malignant or benign, based on the presence of malignant nuclei. Each image was divided into four patches, constituting a bag, to avoid resizing the full-resolution version. The nuclei-specific annotations were not utilized during training, but only during the evaluation, to verify that our model's attention was focused on the nuclei responsible for the image-level label. The dataset was split patient-level, into 80\% training and 20\% test and 5-fold cross-validation was performed for all the models.

The scalability of our method is evaluated on CAM16, which consists of WSI and contours of malignant regions annotated by expert pathologists. We utilized 111 WSI containing malignant regions and 20 WSI containing normal tissue. For training, we sampled $500\times500$ patches from the slides at level 0 magnification and deployed an 80/20 patient-level split. The patches were organized into bags, labeled as malignant if there was an overlap between them and the ground truth contours. The models were trained once, due to the extensive size of the dataset, of over 100,000 patches. 

\noindent\textbf{Model Training}
The evaluated models were trained with cross-entropy loss and optimized with Adam optimizer with a decaying learning rate initialized at 1e\textsuperscript{-4} on an NVIDIA Titan XP GPU. The models, in all cases except for Attention-based Deep Multiple Instance Learning (ADMIL)~\cite{icmlmil}, were initialized with ImageNet weights. The training and visualization framework was implemented in PyTorch\footnote[2]{The code will be released upon acceptance to promote scientific reproducibility.}. The evaluation metric reported for the quantitative comparison of the proposed against the baseline methods was the Classification Accuracy, both patch-wise following single-instance approaches (SIL) and on bag-level (MIL). 

\noindent\textbf{Quantitative Evaluation}
In order to showcase the improvements offered by our method over the original BagNet, we performed ablative testing.
We evaluated models adding each branch individually and combined, as proposed. Furthermore, we compared our method against a variety of baselines to highlight its performance gain. Specifically, we compared our model with the recent work of Ilse et al.~\cite{icmlmil}, utilizing two variations: ADMW is an adaptation of the aforementioned method, ADMIL, trained with $27\times27$ patches taken exhaustively from an image, while ADMP is trained following the convention in the original paper, utilizing only the $27\times27$ patches containing nuclei. Moreover, we equipped two popular architectures, namely ResNet-50~\cite{resnet_conf} and DenseNet-161~\cite{huang2017densely}, with MIL and SIL branches, referred to as RN-50 and DN-161 respectively.

\noindent\textbf{Qualitative Evaluation}
A crucial contribution of this work is the detailed, interpretable attention maps achieved by our model. We highlight this by comparing with popular visualization techniques, namely CAM~\cite{zhou2016learning} and GradCAM~\cite{selvaraju2017grad} using features from the last convolutional layers of the models.

\noindent\textbf{Clinical Usability Assessment}
Two expert pathologists were consulted, regarding the comparison of our produced heatmaps with the baseline methods. Additionally, we aimed to investigate the interpretation value of our method and its potential integration to their workflow.

%% file: 4_results.tex
\section{Results and Discussion}

\subsection{Quantitative Results}
The results of the ablative evaluation are reported in Table~\ref{table:ablative}. The proposed method consistently outperforms the original BagNet~\cite{bagnet} and MILBagNet by 2\%-6\%, for CRC both for MIL and SIL-level accuracies. Regarding CAM16, an improvement of 2\% across the board is achieved by the proposed model, validating our hypothesis that both MIL and SIL branches are required.

Regarding the comparison with baseline models as reported in Table~\ref{table:baseline}, the proposed method outperforms both ADMW and ADMP by 2\%-7\%, for both CRC and CAM16. Furthermore, an improvement of 2\%-5\% was achieved over the traditional RN-50, equipped with MIL and SIL branches, highlighting the contribution of the smaller receptive field in our model. The higher accuracy of our method is consistent when comparing with DN-161 and ranges between 2\% and 3\%.


\begin{figure}[t]
\centering
\begin{minipage}{\textwidth}
\begin{minipage}[t]{0.7\textwidth}
\centering
    \includegraphics[width=\textwidth]{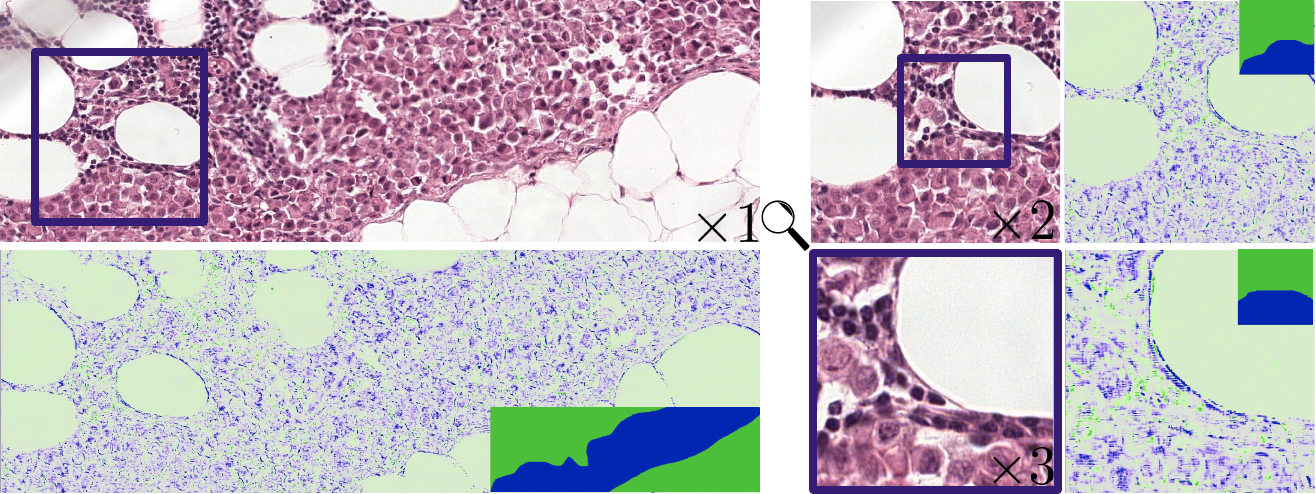}
  \end{minipage}
      \begin{minipage}[t]{0.29\textwidth}
  \vspace{-3.5cm}
      \captionof{figure}{Logit Heatmaps are scalable to WSI maintaining fine grained attention in different levels of magnification. 
      GT masks are shown on the corner of the heatmaps for reference.
      }
      \label{fig:zoom}
    \end{minipage}
  \end{minipage}
\end{figure}

\subsection{Qualitative Results}

\noindent\textbf{Effect of small receptive field}
As can be seen in
Fig.~\ref{fig:crccompare}, 
\begin{wrapfigure}[14]{r}{0.27\textwidth}
    \vspace{-0.6cm}
    \includegraphics[width=0.27\textwidth]{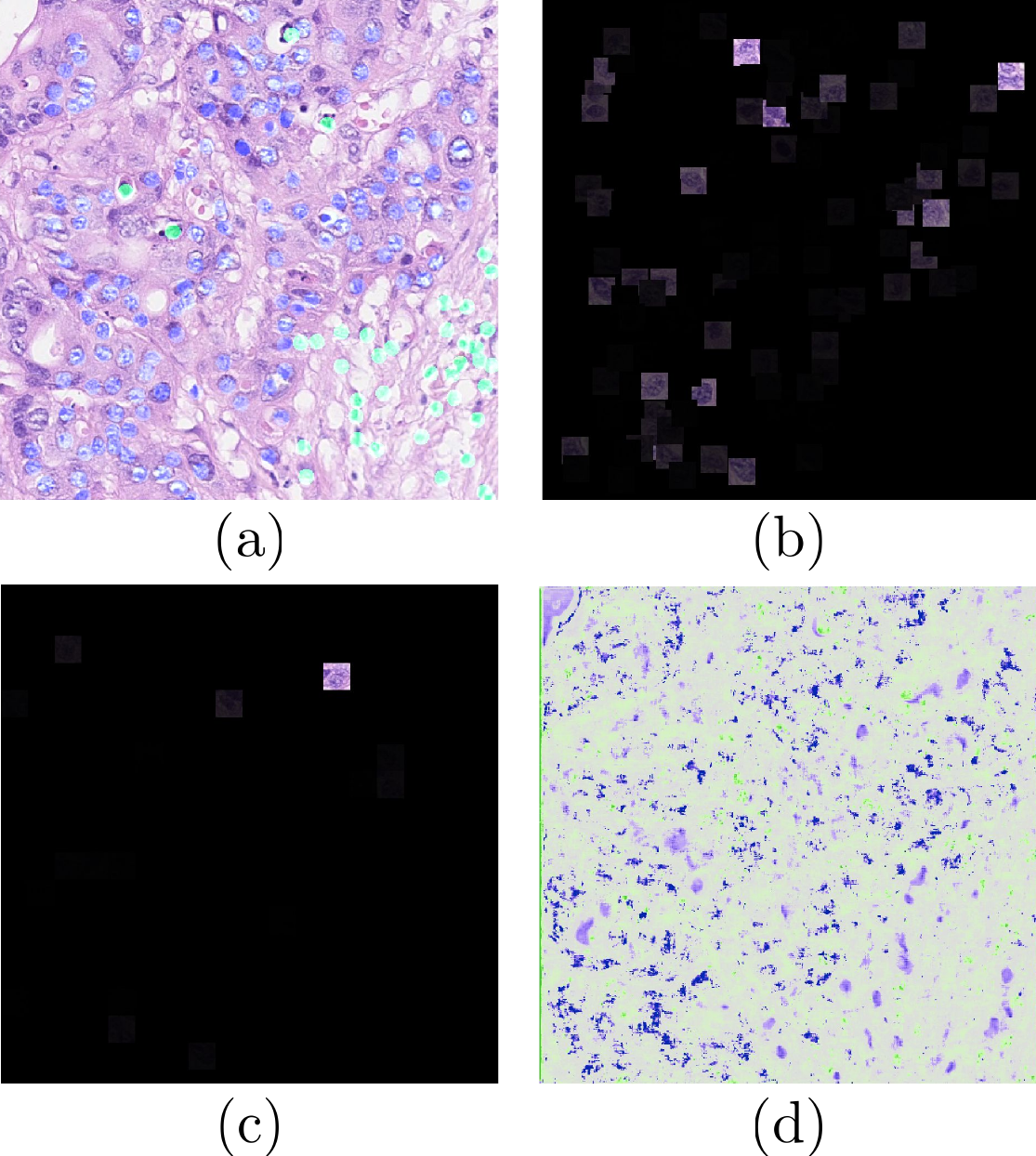}
    \caption{\scriptsize{Visual Comparison with~\cite{icmlmil} on CRC. (a)-Label, (b)-ADMP, (c)-ADMW, (d)-Proposed}}
    \label{fig:vsadmil}
\end{wrapfigure}
CAM and GradCAM were visualized for models trained on both the resized versions of entire images and patch-wise. 
It can be observed in both cases that models trained without a limited receptive field focus their attention on global structures within the images and are not detailed enough for WSI. Furthermore, the tiling effect disrupts the continuity of the attention map when processed patch-wise. 

On the contrary, when CAM and GradCAM are utilized to visualize the attention of the proposed model, the level of detail in the map is much higher and the comparison with the given labels showcases high activations around the cells annotated as malignant.
Moreover, Fig.~\ref{fig:crccompare} also highlights that the logit heatmap visualization achieves the most fine-grained visualization across the baselines.
The architectural limitations of ADMIL (Lack of SIL branch and nuclei-level annotations for CAM16) prevented several experiments denoted by N/A in~Table~\ref{table:baseline}.
\begin{figure}[t]
\centering
\begin{minipage}{\textwidth}
  \begin{minipage}[t]{0.35\textwidth}
  \vspace{-1.05cm}
 \captionof{table}{Ablative evaluation. Average accuracy and std are reported for CRC and accuracy for Camelyon16.}
 \label{table:ablative}
    \end{minipage}
      \hfill
  \begin{minipage}[t]{0.65\textwidth}
\centering

\resizebox{\textwidth}{!}{
\begin{tabular}{@{}lcccccccc@{}}
\multicolumn{2}{c}{} & \phantom{abc} & \multicolumn{1}{c}{BagNet} & \phantom{abc} & \multicolumn{1}{c}{MILBagNet} & \phantom{abc} & \multicolumn{1}{c}{{Proposed}} \\ \toprule
  \parbox[t]{3mm}{\multirow{2}{*}{\rotatebox[origin=c]{90}{\small{{CRC}}}}} & \scriptsize{SIL} & & 87.25 $\pm$ 1.86 & & 83.75 $\pm$ 1.53 & & \textbf{89.59} $\mathbf{\pm}$ \textbf{3.03} \\ &
\scriptsize{MIL} & & 92.00 $\pm$ 2.74 & & 94.00 $\pm$ 2.24 & & \textbf{98.00} $\mathbf{\pm}$ \textbf{2.74} \\ \toprule
   \parbox[t]{3mm}{\multirow{2}{*}{\rotatebox[origin=c]{90}{\tiny{CAM16}}}} & \scriptsize{SIL} & & 82.60 & & 83.37 & & \textbf{85.35} \\ &
 \scriptsize{MIL} & & 84.57 & & 85.22 & & \textbf{86.91} \\
\bottomrule
\end{tabular}
}
  \end{minipage}
  \end{minipage}
\end{figure}

\begin{figure}[t]
\centering
\resizebox{\textwidth}{!}{
\begin{tabular}{@{}lccccccccccc@{}}
\multicolumn{2}{c}{} & \phantom{abc} & \multicolumn{1}{c}{ADMW} & \phantom{abc} & \multicolumn{1}{c}{ADMP} & \phantom{abc} & \multicolumn{1}{c}{RN50} & \phantom{abc} & \multicolumn{1}{c}{DN161} & \phantom{abc} & \multicolumn{1}{c}{{Proposed}} \\ \toprule
\parbox[t]{3mm}{\multirow{2}{*}{\rotatebox[origin=c]{90}{\scriptsize{CRC}}}} & \scriptsize{SIL} & & \scriptsize{N/A} & & \scriptsize{N/A} & & 86.00 $\pm$ 2.56 & & 87.00 $\pm$ 0.68 & & \textbf{89.85} $\mathbf{\pm}$ \textbf{3.03} \\ &
\scriptsize{MIL} & & 88.00 $\pm$ 11.23 & & 91.84 $\pm$ 6.19 & & 93.00 $\pm$ 2.74 & & 95.00 $\pm$ 3.54 & & \textbf{98.00} $\mathbf{\pm}$ \textbf{2.74} \\ \toprule
  \parbox[t]{3mm}{\multirow{2}{*}{\rotatebox[origin=c]{90}{\tiny{CAM16}}}} & \scriptsize{SIL} & & \scriptsize{N/A} & & \scriptsize{N/A} & & 83.76 & & 83.34 & & \textbf{85.35} \\ &
 \scriptsize{MIL} & & 82.75 & & \scriptsize{N/A} & & 84.91 & & 84.96 & & \textbf{86.91} \\ 
\bottomrule
\end{tabular}
}
\captionof{table}{Comparison of baselines with proposed methods. MIL- and SIL- Level accuracies are reported.}
\label{table:baseline}
\end{figure}

\noindent\textbf{Comparison with ADMIL}
Fig.~\ref{fig:vsadmil} compares the performance of the proposed method with the attention maps produced by~\cite{icmlmil}. In the case of ADMIL, the $27\times27$ patches with the highest attention within the bags are visualized with different intensities according to their importance. The attention maps of ADMIL shown in Fig.\ref{fig:vsadmil}(b) and (c) are focused on limited patches from the image and are overall less interpretable compared to our model's attention visualization (Fig.~\ref{fig:vsadmil}~(d)). 


\noindent\textbf{Scalability to WSI}
A crucial advantage of our method is highlighted in Fig.~\ref{fig:zoom}, where a region of a WSI from CAM16 is visualized in magnification level 0. At first glance, our logit heatmaps seem to focus on all the regions of the image that are densely packed with cells. However, after looking at the magnified image, marked by $\times2$ and $\times3$, it is clear that the heatmaps maintain their level of detail, and can scale further to cell structures without having been explicitly trained for that.

\subsection{Clinical Discussion}
Initially, the two consulted expert pathologists were asked to select their preferred method for visualization between CAM, GradCAM and the proposed logit heatmaps by comparing images from each dataset. They commented positively on the fine-grained details of our visualization and agreed that they could interpret the model's decision easily, because of its fine level of detail.

Moreover, the physicians indicated that the proposed method could be utilized successfully as an initial suggestion for an automatic annotation tool, to aid with cumbersome tasks such as tissue microarray annotation. Another recommended application was the assistance of training pathologists utilizing CAD-based systems in their workflow, due to the high level of interpretability. 

%% file: 5_conclusion.tex
\section{Conclusion}

In this paper, a novel interpretation method tailored to WSI was proposed, consisting of a MIL model trained on local features and a fine-grained logit heatmap visualization. Our method was thoroughly evaluated on two challenging, public datasets and outperformed existing approaches, both in the quantitative, and qualitative evaluation. Two expert pathologists verified its potential for clinical integration and interpretation value. Future work includes leveraging our method as an initial proposal for automatic annotation tools and further increasing our interpretability with automatic captioning.

